\pdfoutput=1

\documentclass[11pt]{article}

\usepackage{emnlp2021}

\usepackage{times}
\usepackage{latexsym}

\usepackage[T1]{fontenc}

\usepackage[utf8]{inputenc}

\usepackage{microtype}

\usepackage{times}
\usepackage{latexsym}

\usepackage{xcolor}
\usepackage{graphicx}
\usepackage{caption}
\usepackage{subcaption}
\usepackage{stmaryrd}
\usepackage{amsmath,amsfonts}
\usepackage{float} 
\usepackage{diagbox}

\usepackage{microtype}
\usepackage{enumitem,comment}
\usepackage{multirow}

\title{SLIM: Explicit Slot-Intent Mapping with BERT for Joint Multi-Intent Detection and Slot Filling}

\author{Fengyu Cai, Wanhao Zhou, Fei Mi \and Boi Faltings \\
        LIA, EPFL, Lausanne, Switzerland \\
        \texttt{\{fengyu.cai,wanhao.zhou,fei.mi,boi.faltings\}@epfl.ch}}

\begin{document}
\maketitle
\begin{abstract}
Utterance-level intent detection and token-level slot filling are two key tasks for natural language understanding (NLU) in task-oriented systems. Most existing approaches assume that only a single intent exists in an utterance. However, there are often multiple intents within an utterance in real-life scenarios. In this paper, we propose a multi-intent NLU framework, called SLIM, to jointly learn multi-intent detection and slot filling based on BERT. To fully exploit the existing annotation data and capture the interactions between slots and intents, SLIM introduces an explicit slot-intent classifier to learn the many-to-one mapping between slots and intents. Empirical results on three public multi-intent datasets demonstrate (1) the superior performance of SLIM compared to the current state-of-the-art for NLU with multiple intents and (2) the benefits obtained from the slot-intent classifier.
\end{abstract}

\section{Introduction}

Natural language understanding (NLU) is an essential task for task-oriented dialog (ToD) systems. It contains two major sub-tasks, \textbf{intent detection (ID)} and \textbf{slot filling (SF)}. Take \textit{``Listen to \underline{Westbam} album \underline{Allergic} on \underline{Google music}"} as an example. The task of ID is to identify the intent (``\textit{PlayMusic}'') of the utterance. SF is a sequence labeling task to predict the slot for each token, which are \texttt{[O, O, B-artist, O, B-album, O, B-service, I-service]} using the ``Inside-outside-begining'' (IOB) tagging format \cite{DBLP:conf/ijcai/ZhangW16a}. Traditional techniques often tackle these two tasks separately \citep{haffner2003optimizing, DBLP:conf/emnlp/Kim14,DBLP:conf/slt/YaoPZYZS14,DBLP:conf/naacl/YangYDHSH16,DBLP:conf/interspeech/Vu16}. Recently, models that learn these two tasks jointly achieve better performance by capturing semantic dependencies between intent and slots \citep{DBLP:conf/interspeech/LiuL16,DBLP:conf/naacl/GooGHHCHC18, DBLP:conf/emnlp/LiuMZZCX19, DBLP:journals/corr/JointBert,DBLP:conf/acl/ZhangLDFY19,DBLP:conf/acl/ENCS19,DBLP:conf/emnlp/QinCLWL19}.

\begin{figure}[t!]
		\centering
		\begin{subfigure}[b]{\columnwidth}
		\includegraphics[width=\textwidth]{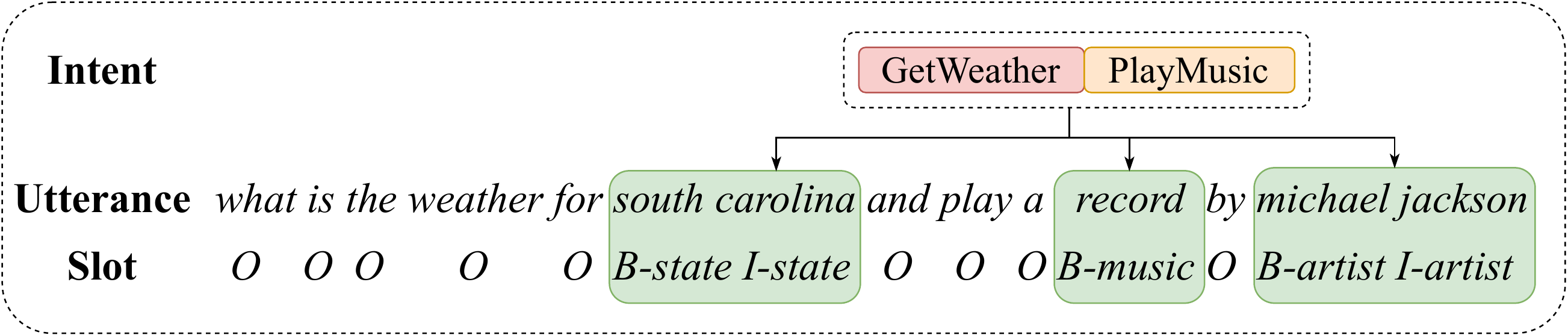}
		\vspace{-1\baselineskip}
		\caption{\citet{DBLP:conf/naacl/GangadharaiahN19,DBLP:conf/emnlp/QinXCL20}: Utterance-level intent is shared by all the slots.}
		\label{fig:example-a}
		\end{subfigure}
		\hfill
		
        \begin{subfigure}[b]{\columnwidth}
		\includegraphics[width=\textwidth]{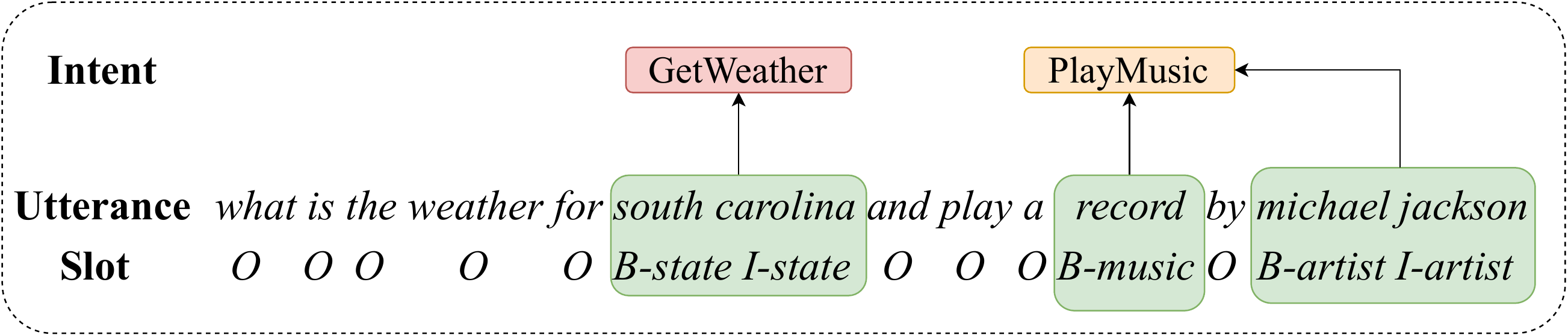}
		\vspace{-1\baselineskip}
		\caption{SLIM: Explicit slot-level intent is used by each slot.}
		\label{fig:example-b}
		\end{subfigure}
		\hfill
		\vspace{-1\baselineskip}
		\caption{Comparison between approaches of utilizing intent information for different slots.}
		\label{fig:example}
\vspace{-1\baselineskip}
\end{figure}

The aforementioned methods assume only a single intent in an utterance. However, multiple intents often exist in one utterance in real-life scenarios \cite{DBLP:conf/iwsds/KimDBWH16,DBLP:conf/naacl/GangadharaiahN19,DBLP:conf/emnlp/QinXCL20,DBLP:conf/lrec/EricGPSAGKGKH20}. 
Regardless of single or multiple intents, the relationship between slots and intent(s) is many-to-one. In single-intent utterances, all slots share the same intent; in multi-intent scenarios, however, different slots may correspond to different intents. 
In Figure \ref{fig:example}, the slot \textit{``south carolina''} corresponds to \textit{GetWeather}, while slots \textit{``record''} and \textit{``michael jackson''} should be linked to \textit{PlayMusic}. Existing multi-intent NLU models \citep{DBLP:conf/naacl/GangadharaiahN19,DBLP:conf/emnlp/QinXCL20} inherit the single-intent pattern, i.e., the
utterance-level intent label distribution is shared by all the slots (Figure \ref{fig:example-a}). Therefore, the relationship between slots and intents are not utilized even though they are \textbf{already} annotated in most NLU datasets.

To this end, we propose SLIM (\textbf{SL}ot-\textbf{I}ntent \textbf{M}apping) that fully exploits the annotations to explicitly captures the mapping between slots and intents. We add a slot-intent classifier to predict the intent label for each slot of an utterance (Figure \ref{fig:example-b}) on top of the state-of-the-art pre-trained model \cite{DBLP:journals/corr/JointBert} based on BERT \cite{DBLP:conf/naacl/DevlinCLT19} to jointly tackle ID and SF. In experiments, we compare SLIM to a wide range of NLU techniques on two simulated datasets (MixATIS and MixSNIPS \citet{DBLP:conf/emnlp/QinXCL20}) and a real-world dataset (DSTC4 \citet{DBLP:conf/iwsds/KimDBWH16}). Empirical results demonstrate that SLIM achieves better performance compared to the current state-of-the-art \textit{without} using extra annotation data. We analyze and reveal that the explicit slot-intent mapping module indeed helps the model learn faster and better. To our best knowledge, this is the first study to explicitly link slots to their intents for better NLU in task-oriented dialog systems.

\section{Approach}

\begin{figure*}[t!]
		\centering
		\includegraphics[width=0.98\textwidth]{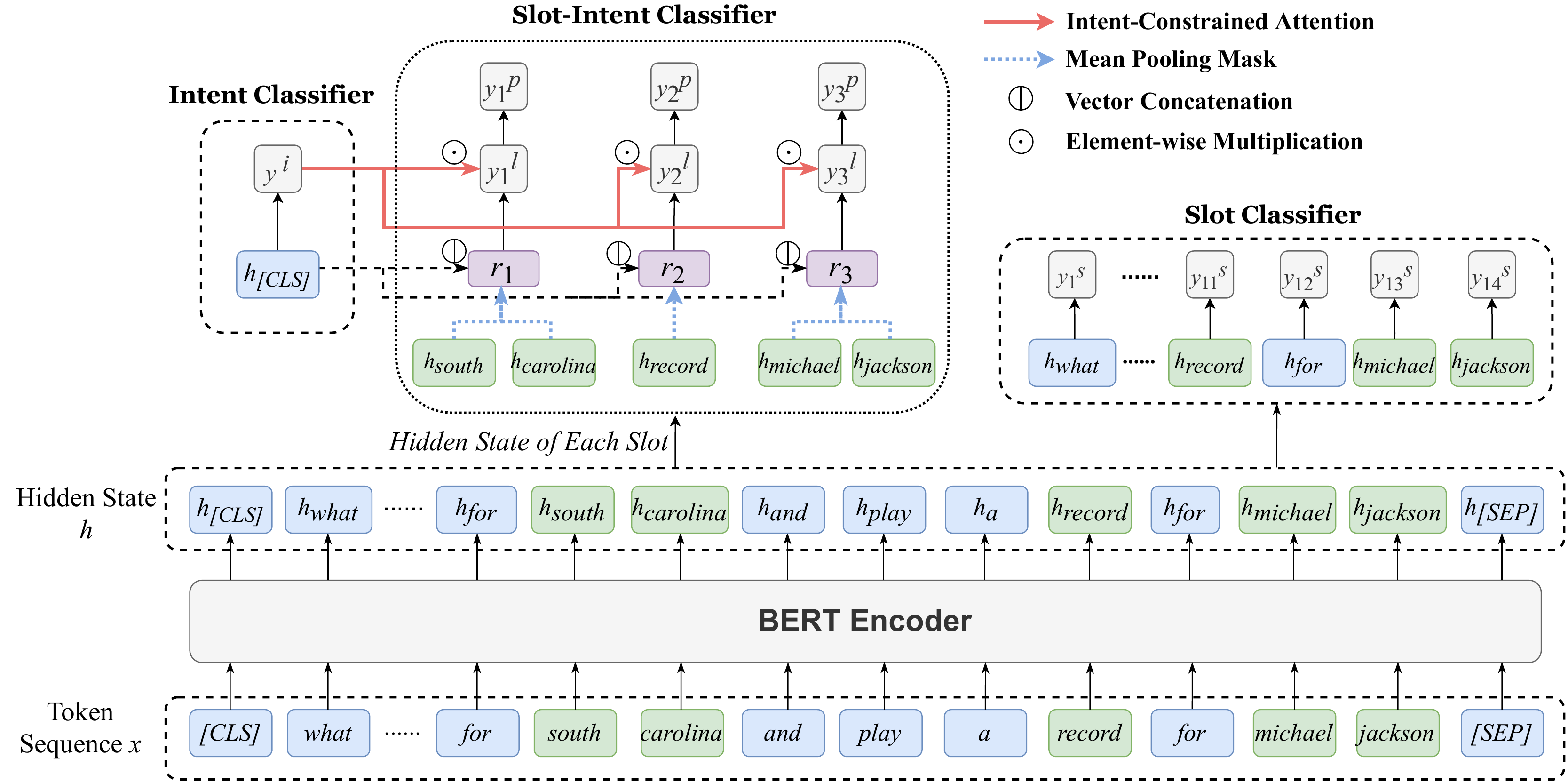}
		\vspace{-0.02\baselineskip}
		\caption{Illustration of our model \textbf{SLIM}. The lower part is the BERT encoder. On the top, from left to right, they are the intent classifier (for ID), the slot-intent classifier, and the slot classifier (for SF) respectively. 
		\label{fig:model_slim}
}
\vspace{-0.07in}
\end{figure*}


\subsection{Problem Setting}
For an input utterance $x$ with token sequence $x = (x_1, ..., x_n)$, the multi-intent NLU task is composed of (1) \emph{utterance-level intent detection}: predict the multi-label intents $I_x \subset I $ of the utterance, where $I$ is the set of possible intents, and (2) \emph{token-level slot filling}: predict the slot label for each token of the input utterance from a set $T$ of possible slots. Different from existing single-intent NLU task, the following two \emph{assumptions} are made:

\begin{enumerate}[label=(\alph*)]
    \item An utterance $x$ has \textbf{at least one} utterance-level intent, i.e, $|I_x| \geq 1$.
    \item \label{assumption:b} Each slot $s_m = \{x_{m_1}, ..., x_{m_j}\}$ is a set of $j$ tokens, and it is mapped to a specific slot-level intent \textbf{belonging to} the utterance-level intents, i.e, $i_m \in I_x$. 
\end{enumerate}

\subsection{Model}
\label{sec:model}
The proposed model (SLIM) contains a shared encoder and three classifiers for different tasks.

\paragraph{Encoder} We use BERT \cite{DBLP:conf/naacl/DevlinCLT19} as token sequence encoder of our model. The utterance is tokenized by standard BERT tokenizer with a special token [CLS] prepended and [SEP] appended. The output $h = (h_{\mathrm{cls}}, h_1, ..., h_n, h_{\mathrm{sep}})$ of BERT's encoder will be utilized for three classifiers, where $h_k \in \mathbb{R}^d$.

\paragraph{Intent Classifier and Slot Classifier} Utterance-level intent detection is accomplished by the intent classifier. To classify the intent $y^i$, we use sigmoid as the activation function after feeding $h_{\mathrm{cls}}$ into a output network by:
\begin{equation}
y^i = Sigmoid(W^i h_\mathrm{cls} + b^i),
\label{eq:intent}
\end{equation}
where $W^i \in \mathbb{R}^{|I|\times d}$, and each dimension of $y^i \in \mathbb{R}^{|I|}$ represents the probability of an intent label.
In the slot filling task, to predict the slot $y^s_k$ at position $k$, we apply softmax after feeding $h_k$ into a separate slot classification network as:
\begin{equation}
    y^s_k = Softmax(W^s h_k + b^s),
    \label{eq:slot}
\end{equation}
where $W^s \in \mathbb{R}^{|T| \times d}$ , and $y^s_k \in \mathbb{R}^{|T|}$ is the slot probability distribution for token $x_k$ with $k \in \{1,\dots, n\}$.
Intent classifier and slot classifier are formulated similarly as \citet{DBLP:journals/corr/JointBert}, except that intent detection is formulated as a multi-label classification task.

\paragraph{Slot-Intent Classifier} To explicitly capture the relation between slots
and intents, we predict the slot-level intent for each slot $s_m$. First, we compute the representation $r_m$ of a slot $s_m=\{x_{m_1}, ..., x_{m_j}\}$ by a \textit{mean pooling} of the token representations in this slot by: $r_m = 1 / j \sum_{i=1}^{j} h_{m_i}$. Afterward, we concatenate the global utterance representation $h_{\mathrm{cls}}$ with $r_m$ and compute an \textbf{unconstrained} slot-intent prediction $y^l_m$ as:
\begin{equation}
    y^l_m = Softmax(W^l [h_{\mathrm{cls}} | r_m] + b^l),
\end{equation}
 where $W^l \in \mathbb{R}^{|I| \times 2d}$, and $|$ indicates vector concatenation.
 To better align the above slot-intent prediction with the predicted utterance intent~(\hyperref[assumption:b]{assumption (b)}), we propose an \emph{intent-constrained attention}. It computes a final \textbf{constrained} slot-intent prediction $y^p_m$ by an element-wise multiplication between the utterance-level intent prediction $y^i$ and unconstrained slot-intent prediction $y^l_m$ as:
 \begin{equation}
    y^p_m = y^i \odot y^l_m
\end{equation}

\subsection{Training Objective}
The global objective of SLIM is to maximize $p(y^i, y^s, y^p | x)$, and it can be decomposed for tokens inside and outside of slots as:

\begin{align}
& \prod_{m} \underbrace{p(y^i, y^s_{s_m}, y_m^p | x)}_{\text{inside each slot}} \underbrace{\prod_{ j \notin \cup_{m}s_m} p(y^i, y_{j}^s | x)}_{\Gamma:~\text{outside the slots}}\\
     = ~ & \prod_{m} p(y^i, y^s_{s_m} | x)\cdot p(y_m^p | y^i, y^s_{s_m}, x) \cdot \Gamma \\ 
   \propto ~ & p(y^i|x) \prod_{k = 1}^{n} p(y_k^s|x) \prod_{m} p(y_m^p | y^i, y^s, x) \\
   = ~ & \underbrace{p(y^i|x)}_{\text{ID}} 
   \underbrace{\prod_{k = 1}^{n} p(y_k^s|x)}_{\text{SF}}
   \underbrace{\prod_{m} p(y_m^p | y^i, y_{s_m}^s, x)}_{\text{Slot-Intent Classification}} \label{eqn:global}
\end{align}

In equation \ref{eqn:global}, the first two terms are the objectives of ID and SF, and the last term is the objective of slot-intent mapping. ID is trained with binary cross-entropy loss for multi-intent detection, while the two other terms are trained with regular cross-entropy loss. Because the slot-intent prediction $y^p_m$ is conditional on the predicted utterance intent $y^i$ and slot label $y_{s_m}^s$ of the slot, parameters for ID and SF will also be updated when training the slot-intent classifier.  
The loss of SLIM is a weighted sum of losses from these three classifiers. An overview model pipeline of SLIM is illustrated in Figure \ref{fig:model_slim}.

\section{Experiment and Analysis}
\begin{table*}[!t]
\centering
\small
\begin{tabular}{l|c|c|c|c|c|c}
\hline
\multicolumn{1}{c|}{\multirow{2}{*}{\textbf{Model}}} & \multicolumn{3}{c|}{\textbf{MixATIS}}             & \multicolumn{3}{c}{\textbf{MixSNIPS}}            \\ \cline{2-7} 
\multicolumn{1}{c|}{}                       & \textbf{Slot F1} & \textbf{Intent Acc} & \textbf{SeFr Acc} & \textbf{Slot F1} & \textbf{Intent Acc} & \textbf{SeFr Acc} \\ \hline
    
Bi-Model$^{\dag}$~\citep{DBLP:conf/naacl/WangSJ18} & 85.5 & 72.3 & 39.1 & 86.8 & 95.3 & 53.9 \\ 
SF-ID$^{\dag}$~\citep{DBLP:conf/acl/ENCS19} & 87.7 & 63.7 &  36.2 & 89.6 & 96.3 & 59.3 \\ 
Stack-Propagation$^{\dag \S}$~\citep{DBLP:conf/emnlp/QinCLWL19} & 86.6 & 76.0 & 42.8 & 93.9 & 96.4 & 75.5 \\ 
Joint Multiple ID-SF$^{\ddag}$~\citeyearpar{DBLP:conf/naacl/GangadharaiahN19} & 87.5 & 73.1 & 38.1 & 91.0 &  95.7 &  66.6 \\
AGIF$^{\ddag \S}$~\citep{DBLP:conf/emnlp/QinXCL20} & 88.1 & 75.8 & 44.5 & 94.5 & 96.5 & 76.4 \\ \hline
SLIM (w/o slot-intent classifier) & 85.6 & 77.1  & 46.3  & 96.2  & 96.8  & 82.3  \\
SLIM (w/o intent-constrained attention) & 87.2 & 75.6 & 46.4  & 96.5  & 96.0 & 83.6 \\
SLIM & \textbf{88.5} & \textbf{78.3} & \textbf{47.6} & \textbf{96.5} & \textbf{97.2} &  \textbf{84.0} \\    \hline
\end{tabular}
\caption{Results on two multi-intent datasets. $^\dag$ or $^\ddag$ denotes single-intent or multi-intent model respectively. Results of models with $^\dag$ and $^\ddag$ are taken from \citet{DBLP:conf/emnlp/QinXCL20}. Models with $^\S$ indicate the previous state-of-the-art solutions. \textbf{Bold} numbers are the best results in each column.}
\label{tab:table1}
\end{table*}


\begin{table}[!t]
\centering
\small
\begin{tabular}{l|c|c|c}
\hline
\multicolumn{1}{c|}{\multirow{2}{*}{\textbf{Model}}} & \multicolumn{3}{c}{\textbf{DSTC4}}               \\ \cline{2-4} 
\multicolumn{1}{c|}{}                       & \textbf{Slot F1} & \textbf{Intent Acc} & \textbf{SeFr Acc} \\ \hline
Stack-Propagation$^\dag$ & 56.4 & 35.8 & 20.7 \\
AGIF$^\ddag$ & 57.6 & 33.0 & 19.4 \\ 
SLIM & \textbf{61.1} & \textbf{36.7} & \textbf{21.3} \\ \hline

\end{tabular}
\caption{Results on DSTC4. $^\dag$ or $^\ddag$ denotes single-intent or multi-intent model respectively. \textbf{Bold} numbers are the best results in each column.}
\label{tab:table2}
\vspace{-0.07in}
\end{table}
\paragraph{Dataset} We evaluate our approach on three multi-intent SLU datasets. \textbf{MixSNIPS}  \citep{DBLP:conf/emnlp/QinXCL20} contains 39,776/2,198/2,199 utterances for train/validation/test. It is created based on the Snips personal voice assistant \citep{DBLP:journals/corr/abs-1805-10190}. \textbf{MixATIS}  \citep{DBLP:conf/emnlp/QinXCL20} is constructed from ATIS dataset \citep{DBLP:conf/naacl/HemphillGD90}, containing 13,161/759/828 utterances for train/validation/test. \textbf{DSTC4} \cite{DBLP:conf/iwsds/KimDBWH16} contains multi-intent human-human dialogues with 5,308/2,098/1,865 utterances for train/validation/test. More details on these datasets are provided in Appendix \ref{appendix:data}.

\paragraph{Training Details} We use English uncased BERT-Base model, containing 12 layers, 768 hidden states, and 12 heads. The max sequence length is 50, and training batch size is 32. Hyper-parameters of SLIM are tuned by a randomized search (see Appendix \ref{appendix:hyper}) based on the semantic frame accuracy on validation set. Dropout rate is 0.2 for the output layers of all three classifiers. In Eq. (\ref{eqn:global}), the losses of ID and slot-intent classifier are weighted by 1, and the loss of SF is weighted by 2. We train SLIM for maximum 20 epochs, and the early stop patience is 3 epochs. 
For the slot-intent classifier, we use slots provided by ground truth during training, while the predicted slots are used during inference.

\paragraph{Baselines} We compare SLIM with both single-intent and multi-intent models. When evaluating single-intent models on the multi-intent task, we follow \citet{DBLP:conf/naacl/GangadharaiahN19, DBLP:conf/emnlp/QinXCL20} to concatenate multiple intents with `\#' into a single intent. On MixATIS and MixSNIPS, we compare previous single-intent models: Bi-Model \citep{DBLP:conf/naacl/WangSJ18}, SF-ID \citep{DBLP:conf/acl/ENCS19}, Stack-Propagation \citep{DBLP:conf/emnlp/QinCLWL19} and recent multi-intent models: Joint Multiple ID-SF \citep{DBLP:conf/naacl/GangadharaiahN19}, AGIF \citep{ DBLP:conf/emnlp/QinXCL20}. On DSTC4, we compare SLIM with state-of-the-art single-intent model Stack-Propagation \citep{DBLP:conf/emnlp/QinCLWL19} and multi-intent model AGIF \citep{DBLP:conf/emnlp/QinXCL20}.

\paragraph{Overall Results} We evaluate the performance of different methods using three metrics: F1  score of slot filling (Slot F1), accuracy of intent detection (Intent Acc), and semantic frame accuracy (SeFr Acc) as in \citet{DBLP:conf/naacl/GangadharaiahN19} and \citet{DBLP:conf/emnlp/QinXCL20}. ``SeFr Acc'' considers the prediction of an utterance to be correct when slots and intents are all accurate.
Table \ref{tab:table1} summarizes different models' performances on MixATIS and MixSNIPS. We observe that SLIM outperforms other baselines w.r.t. all three metrics, especially on the semantic frame accuracy. On Slot F1 and Intent Acc, SLIM outperforms the previous start-of-the-art (AGIF) by 0.4\%-2.5\%. More importantly, SLIM improves the semantic frame accuracy compared to AGIF by 3.1\% and 7.6\% on two datasets respectively. 
This result indicates that SLIM effectively improves the joint correctness for predicting intents and slots for understanding an utterance. 
Table \ref{tab:table2} summarizes the results on the real-world DSTC4 dataset, and similar performance patterns can be observed as in Table \ref{tab:table1}. Moreover, SLIM also notably improves Slot F1 on this dataset (3.5\% and 4.7\% gain over the two baselines), indicating that SLIM can also strengthen token-level prediction in addition to the large utterance-level improvements in Table \ref{tab:table1}.

\begin{figure}[t!]
		\centering
		\includegraphics[width=\columnwidth]{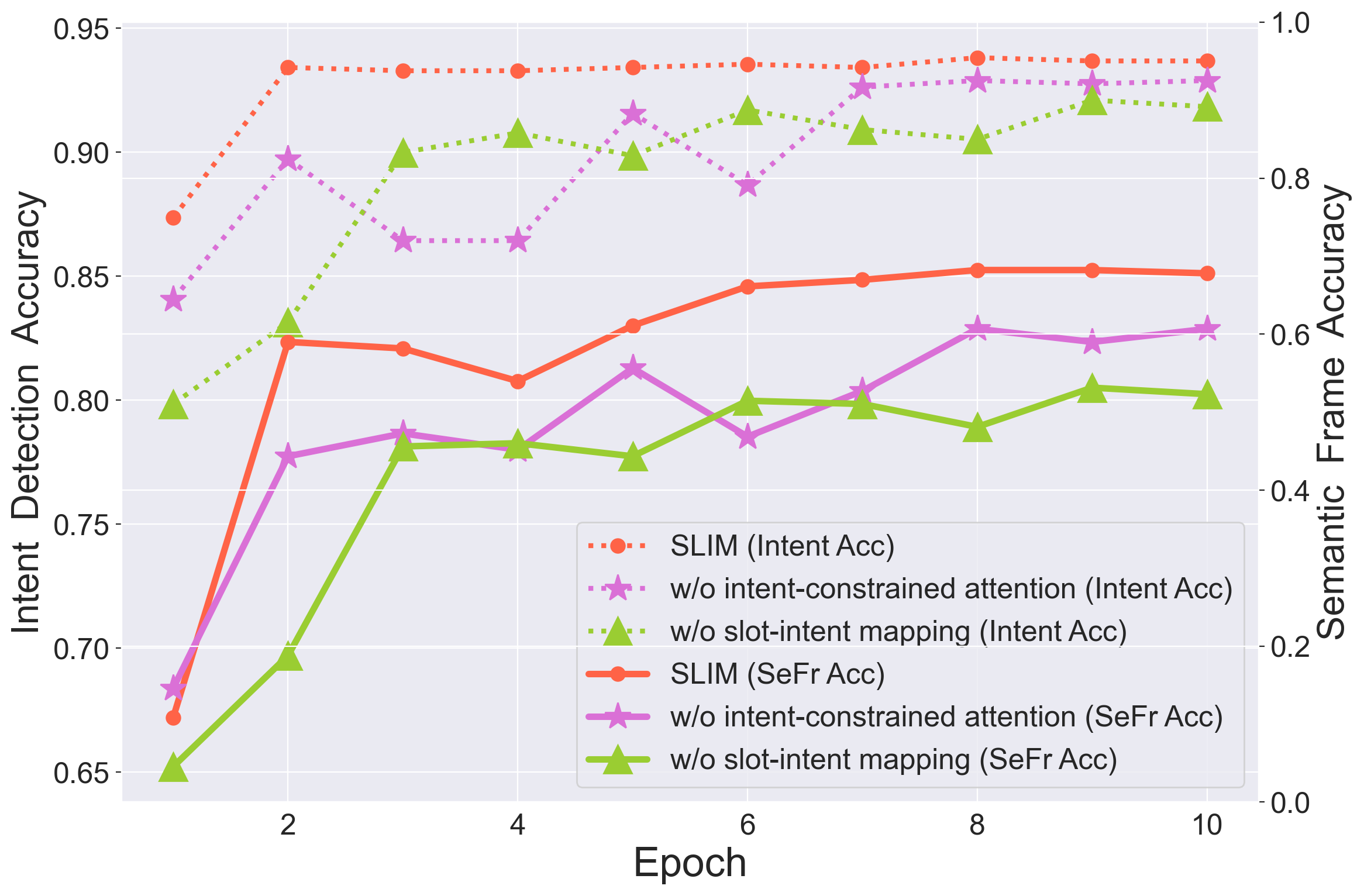}
		\caption{Intent Acc and SeFr Acc on MixATIS validation dataset after each training epoch.}
		\label{fig:compare}
		\vspace{-0.5\baselineskip}
\end{figure}

\paragraph{Ablation Study} We compare SLIM with two simplified versions, \textit{w/o slot-intent classifier}~\footnote{This version degenerates to \citet{DBLP:journals/corr/JointBert} as mentioned in Sec. \ref{sec:model}} and \textit{w/o intent-constrained attention}, to analyze how the slot-intent classifier and its intent-constrained attention affect the performance and the training process. 
From Table \ref{tab:table1} (Bottom), we can see that dropping these two components impairs the performance of SLIM. Besides the lower semantic frame accuracy, dropping slot-intent classifier mainly degrades Slot F1, and dropping intent-constrained attention mainly degrades Intent Acc.
Furthermore, we plot the learning curves of these three methods on MixATIS's validation set in Figure \ref{fig:compare}. We can observe that SLIM converges \textit{faster and better} compared to the two simplified versions, which demonstrates the benefits of the slot-intent classifier with intent-constrained attention.

\section{Conclusion}
In this paper, we propose SLIM for multi-intent NLU in task-oriented dialog systems. SLIM explicitly utilizes the relation between slots and intents by mapping slots to the corresponding intent. 
Experimental results on three public datasets show that SLIM outperforms the previous state-of-the-art NLU models. 
Our findings may inspire future studies to better exploit the relationship between slots and intent for complicated NLU scenarios in task-oriented dialog systems.


\bibliography{emnlp2021}
\bibliographystyle{acl_natbib}
\begin{figure*}[t!]
		\centering
		\includegraphics[width=1.98\columnwidth]{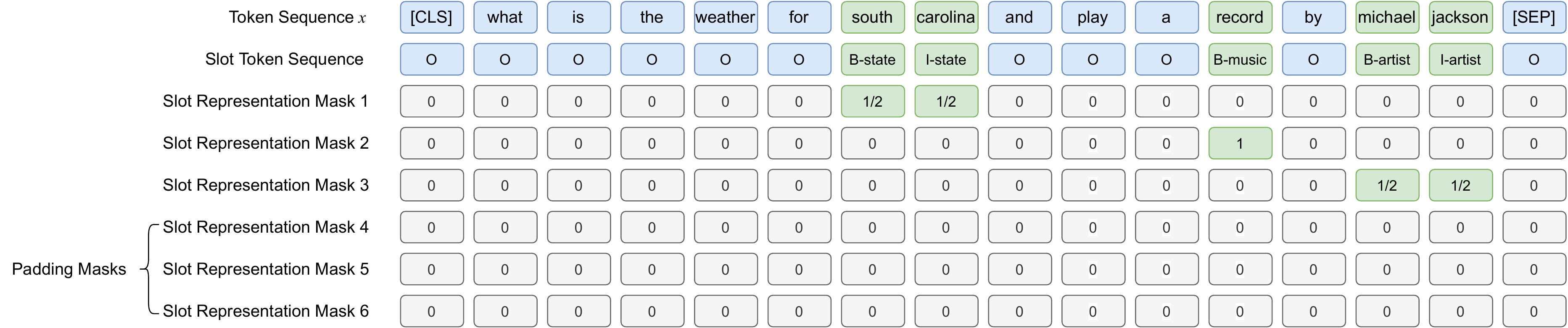}
		\caption{
		\label{fig:slot} Illustration of our slot representation mechanism. In the example sentence above, we have in three actual slots and three additional padding slots. The representation $r_m$ of $m$-th slot, where $m \in \{1,\dots, 6\}$, is the average of hidden states of tokens inside the slot.
		}
\end{figure*}
\newpage
\appendix

\part*{ Appendix}

\section{Reproducibility Checklist}
\label{appendix:repro}
\subsection{Computing Infrastructure and Computation Time}
All experiments are conducted using a single GeForce GTX TITAN X GPU. When training for 10 epochs, time costs are approximately 1 hour, 45 minutes and 30 minutes on MixSNIPS, MixATIS, and DSTC4 respectively.

\subsection{Number of Parameters}
SLIM includes one BERT encoder and three classifiers. Compared with BERT, the parameter size of three output networks is much smaller and is dependent on the number of intent labels ($|I|$) and slot labels ($|T|$) in the target dataset. Therefore, the number of parameters in SLIM is marginally larger than BERT, which is around 110 million.

\subsection{Hyper-parameter Search}
\label{appendix:hyper}
\begin{table}[!t]
\centering
\resizebox{\columnwidth}{!}{
\begin{tabular}{l|l}
\hline
\textbf{Hyper-parameter}  & \textbf{Search Range} \\ \hline
Dropout Rate &  \{0, 0.1, \textbf{0.2}, 0.3, 0.4\}      \\ 
Learning Rate  &  \{1e-5, \textbf{5e-5}, 1e-4, 5e-4\}     \\ 
Loss Weight of Intent Classifier    &  \{0.5, \textbf{1}, 2\}        \\ 
Loss Weight of Slot classifier         &  \{0.5, 1, \textbf{2}\}   \\ 
Loss Weight of Slot-Intent Classifier  &   \{0.5, \textbf{1}, 2\} \\ \hline       
\end{tabular}
}
\caption{\label{tab:hyperparam} Hyper-parameter search range of our proposed \textbf{SLIM} model. \textbf{Bold} numbers indicate our choice of hyper-parameters.}
\end{table}
In total, we have 5 hyper-parameters to configure: dropout rate, learning rate, the weight of losses for intent classifier, slot classifier and slot-intent classifier. We randomized search with 30 trials for the best setting to maximize the semantic frame accuracy on validation set. Detailed search range and our choice of hyper-parameters are given in Table \ref{tab:hyperparam}.

\subsection{Detailed Dataset Specifications}
\label{appendix:data}
Table \ref{tab:nointent} summarizes the number of intent labels and slot labels in the training set of MixSNIPS, MixATIS, and DSTC4. Table \ref{tab:nosentence} reports the statistics of number of intents in utterances. For DSTC4, we split the training and validation data, because the original test data are no longer accessible.

\section{Implementation Details on Slot Representation}
Since the number of slots varies between utterances, to keep the SLIM model running in batch mode, we pad the number of slots in each utterance to 6. We observe that 6 is the 99-th percentile of the number of slots in an utterance in real-world multi-intent dataset DSTC4. We pad slots in each utterance to 6 in order to strike a balance between the memory consumption and coverage of most slots in utterances. An illustrative example is given in Figure \ref{fig:slot}. Each mask extracts a different slot in the utterance. The actual number of slots in this utterance is three, and we have three more padding masks with all zeros.

\begin{table}[!t]
\centering
\small
\begin{tabular}{l|c|c}
\hline
\textbf{Dataset}  & \textbf{\# of Intent Labels} & \textbf{\# of Slot Labels} \\ \hline
MixSNIPS & 7                       & 71                    \\ 
MixATIS  & 21                      & 118                   \\ 
DSTC4    & 20                      & 90                     \\ \hline
\end{tabular}
\caption{\label{tab:nointent} Summary of the number ($|I|$) of intent labels and the number ($|T|$) of slot labels in MixSNIPS, MixATIS and DSTC4.}
\end{table}
\begin{table}[!t]
\centering
\resizebox{\columnwidth}{!}{
\begin{tabular}{l|c|c|c|c}
\hline
\diagbox{\textbf{Dataset}}{\textbf{\# of Sentences}}{\textbf{\# of Intent(s)}} & \textbf{1} & \textbf{2} & \textbf{3}& \textbf{4}              \\ \hline
MixSNIPS & 8,277  & 22,499  & 9,000  & -        \\ 
MixATIS  & 1,118 & 8,444  & 3,599  & -            \\ 
DSTC4    & 3,963 & 1,240    & 100  & 5                    \\ \hline
\end{tabular}
}
\caption{\label{tab:nosentence} Summary of number of utterances with different numbers of intents in MixSNIPS, MixATIS and DSTC4.}
\label{tab:table3}
\end{table}




\end{document}